\documentclass[runningheads]{llncs}

\usepackage[T1]{fontenc}
\usepackage{graphicx,verbatim}
\usepackage{multirow}
\usepackage{amsmath}
\usepackage{bbm}

\begin{document}
\title{Learning to Defer with Guidance on Real World Medical Data}
\author{Emma Sun \and
Joshua Strong \and
Alison Noble}
\authorrunning{E. Sun et al.}
\institute{Department of Engineering Science, University of Oxford, UK\\
Correspondence: \texttt{emma.sun@eng.ox.ac.uk}
}
  
\maketitle             
\begin{center}
    \small{Preprint version.}
\end{center}
\begin{abstract}
Medical image interpretation is high-volume and time-consuming, and while AI interpretation can reduce workload, fully autonomous deployment carries potential safety concerns and low specificity may in practice lead to increased clinician workload. Learning to Defer (L2D) addresses this by selectively routing cases between autonomous prediction and human experts by learning from input features and AI model and human performance. While theoretical guarantees have been proven for L2D, its performance has not been validated on real-world medical datasets with human reader annotations. 
We evaluate the predictor-rejector formulation of two-stage L2D, where the AI predictor model is fixed and separate from the trainable routing or rejector model, on Collab-CXR, a multilabel chest X-ray dataset with multiple human annotations per case. This is the first work to look at L2D in the context of real-world medical imaging data with human annotations. We further introduce a new setup, L2D with Guidance, where the decision space is extended to three choices: predict autonomously, defer to a human expert, or defer to a human expert and provide AI guidance. 
We compare multiple rejector architectures and loss functions, and different input feature availabilities. 
This is reproduced on two larger datasets, VinDr-CXR and CheXpert. 
Our results show that two-stage L2D with Guidance outperforms classic two-stage learning to defer, as well as human-alone, AI-alone and AI-guided human baselines. Notably, this performance is achieved with simpler loss functions compared to formally defined L2D surrogate loss functions in current literature.

\keywords{Learning to Defer \and Human-AI Collaboration \and Computer-Aided Diagnosis.}

\end{abstract}

\section{Introduction} 
Real world medical image interpretation is a high-volume and time-consuming task for clinicians, and delays in image interpretation delay patient care. Various AI models have shown promising performance for some medical image interpretation tasks, but uncertainty around the safety of allowing AI models to make decisions autonomously, or high model sensitivity accompanied by low specificity can result in increased rather than reduced workload for clinicians in practice~\cite{deployAIhealthcare}.

Human-AI Collaboration (HAIC), in particular Learning to Defer (L2D) \cite{l2dmozannar2020consistent}, in healthcare offers a potential approach to support workload reduction. L2D is a supervised learning framework in which a rejector model learns, for each input, whether to output an autonomous decision from the AI predictor, or defer to an external expert, so as to minimise the overall expected system-level cost that accounts for both prediction errors and instance-dependent deferral costs. 
By only deferring to a human expert in cases where it is beneficial, it reduces the number of cases that need review by a human expert while maintaining greater oversight for safety compared to using autonomous AI.  
In the two-stage predictor-rejector setup of L2D, the rejector is trainable but the predictor is treated as fixed \cite{Mao2023Defer}. This flexibility can be useful in medical settings, where AI models (predictors) can be large or proprietary, therefore updating model weights is expensive or forbidden. Two-stage learning to defer is a relatively new addition to the field, and although theoretical guarantees have been proven for the performance, these have not previously been demonstrated with real-world medical-imaging data, with human reader annotations on top of the ground truth annotations. The current standard in the field is to simulate the human reader annotations.

\begin{figure}[t]
    \centering
    \includegraphics[width=0.75\linewidth]{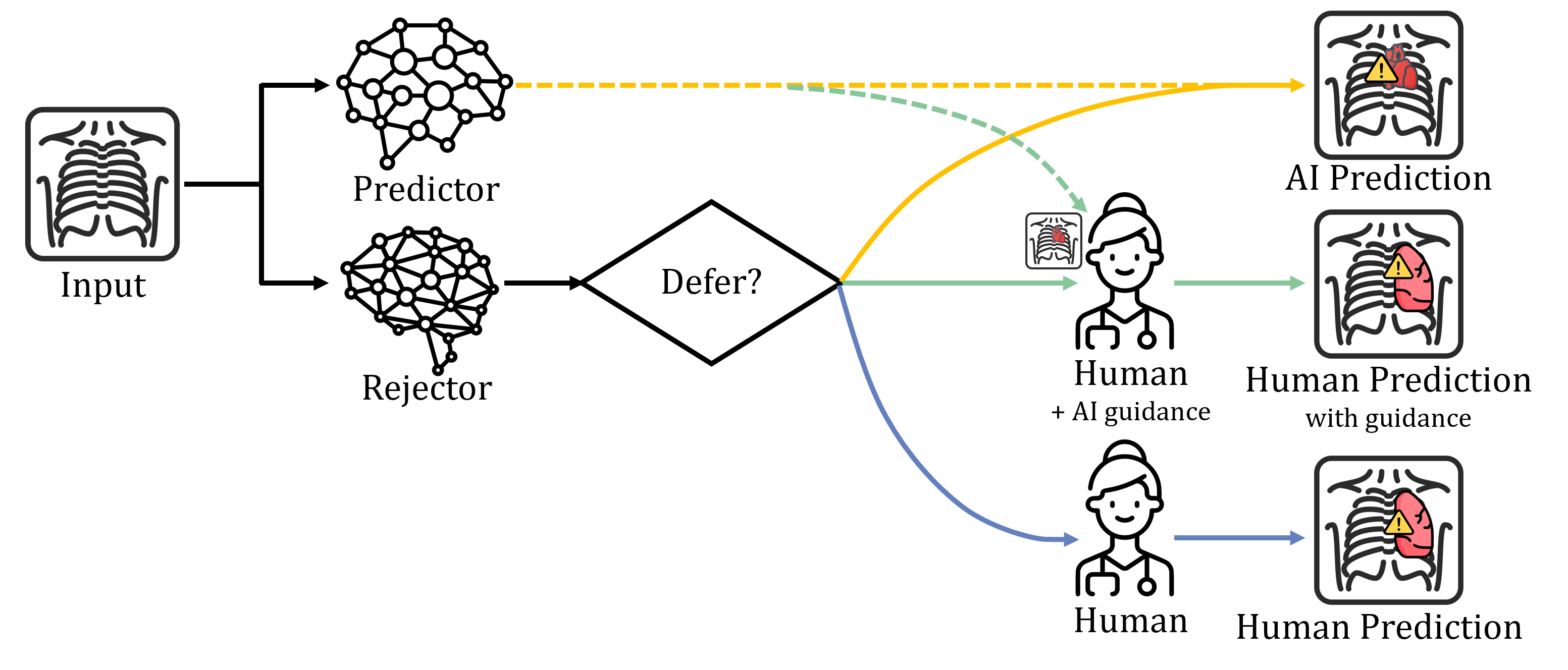}
    \caption{L2D with Guidance setup, where the rejector can choose to output autonomous AI prediction, defer to the AI-guided human expert, or defer to the plain, unguided human expert. Adapted from \cite{Strong_2025}.}
    \label{fig:guideddeferral}
\end{figure}

In this work we propose a new L2D setup: L2D with Guidance (L2D-G). The guidance is produced by an AI model and acts an assistive signal, that may for instance take the form of mask overlays or suggested labels with accompanying confidence levels. Combining AI guidance with L2D creates a system which chooses to defer autonomously, defer to a human expert, or defer to a human expert and provide AI guidance (Fig. \ref{fig:guideddeferral}). AI guidance is not guaranteed to be beneficial to human performance in all cases \cite{Rudolph_2024} – using L2D will mean only showing AI guidance for cases where it will help. AI and different individual clinicians may have different strengths and weaknesses; the core principle of L2D is that the rejector can learn how input features are tied to these performance differences and accordingly select the reader (human, AI, or human with AI assistance) to take advantage of these different strengths. Learning to Defer with Guidance may be particularly useful in the case of out-of-hours care, where clinicians are present but not necessarily those of the optimal speciality. As well as AI producing decisions faster, determining which cases can be safely interpreted by AI or by an available clinician with assistance from AI could allow decisions to be made sooner in the cases where it is safe.

We validate L2D-G on Collab-CXR \cite{CCXR}, a multi-label chest X-ray public dataset of 324 images with multiple human annotations per case, and compare its performance with the classic two-stage L2D setup. To our knowledge, this is the first empirical evaluation of L2D on real-world, human-annotated medical data and the first formulation incorporating conditional AI guidance into the deferral decision. This validation is reproduced on two further datasets, CheXpert and VinDr-CXR. In summary, the contributions of this paper are: 
\begin{itemize}
    \item We introduce the Learning to Defer with Guidance setup,
    \item We evaluate classic L2D for the first time on real-world, human-annotated medical imaging data, and compare the performance of L2D-G and classic L2D. 
\end{itemize}

\section{Method}
\subsubsection{Rejector Models} 
L2D rejectors were implemented, adapted from the traditional L2D action space of \{Defer, Predict\} to extend to the case where human experts are provided with AI guidance. 

\noindent Four candidate L2D rejector models were implemented, two using tree-based classic machine learning models (XGBoost \cite{XGBoost} and Random Forest \cite{RF}), and two using multi-layer perceptrons (MLPs). One MLP model was implemented with a binary cross-entropy loss function, which models expert correctness explicitly, and the other with a formal L2D loss function which models expert correctness implicitly, based on the surrogate loss function from Mao et al. \cite{Mao2023Defer}:
\begin{equation}
L_{h}(r, x,y) = \mathbbm{1}_{h(x) = y} \cdot \ell_2(\bar{r}, x, 0) + \sum_{j=1}^{J} \bar{c}_j(x,y) \cdot \ell_2(\bar{r}, x, j)    
\end{equation}

\noindent where $h$ is the predictor function, $r$ the rejector function, $\bar{r}$ an associated hypothesis where $\bar{r}(x,0)=0$ and $\bar{r}(x,j)=-r_j(x)$. $J$ is the number of experts, 
$\ell_2$ is a standard multi-class loss function, 
and $\bar{c}(x,y) = 1 - c(x,y)$, where $c(x,y)$ is the cost of deferral.

\subsubsection{Input Features}
An ablation study was performed to assess the importance of each of the input features across the different rejectors: image features were either omitted, provided by naive embeddings, or provided by the pathology-specific encoder applied to the naive embeddings. A ResNet model \cite{ResNet} was used to encode each image as a 2048-dimensional feature embedding. It was first pre-trained on a larger (non-overlapping) CXR dataset to increase relevance of encoded features to the task of CXR interpretation. The weight of this encoder were not updated during training, for computational efficiency. 
A pathology-specific encoder was then included to encode the image features according to the pathology being queried, to enable the rejector to focus on the most relevant image features for each pathology accordingly.
This encoder mapped the image features extracted by the ResNet encoder to the embedding space, processed these embedded features with further fully connected layers, and created label-specific embeddings with a separate linear head per pathology label. The weights of this pathology-specific encoder were updated as part of rejector training to allow fine-tuning to the dataset.

\noindent The expert predictions, with and without AI guidance, and the output of the AI tool from the dataset are treated as the outputs of two experts and the predictor respectively. These predictor outputs were not updated during training, as per the two-stage setting of L2D \cite{Mao2023Defer}.

\section{Experiments}
\subsection{Datasets}
\subsubsection{Collab-CXR}
Experiments were performed on data from an public dataset, Collab-CXR \cite{CCXR}, the only suitable dataset which is publicly available. Collab-CXR provides annotations by up to 10 radiologists, with and without AI assistance by CheXbert labeller \cite{cheXbert}, on the probability of the prevalence of 106 pathologies for 324 chest X-ray (CXR) images.

However, the Collab-CXR study design meant that the images were read by an inconsistent combination of the radiologists. This is not suitable for training and testing an L2D system, as the rejector implicitly learns the performance of each expert during training, and accordingly for a naive implementation would be forced to train 10 different rejectors (one for each radiologist), each on a much smaller dataset than the original. It was therefore necessary to pool the predictions of the radiologists for each setting to get a single meta-expert probability for each pathology-image combination. While this pooling `smooths' over differences in performance across individual readers, it enables training on the whole dataset and preserves differences between human-alone and AI-guided humans performance, and is more realistically-grounded in human performance characteristics compared to synthetic annotations that are the current standard in L2D literature.

\begin{figure}
    \centering
    \includegraphics[width=0.8\linewidth]{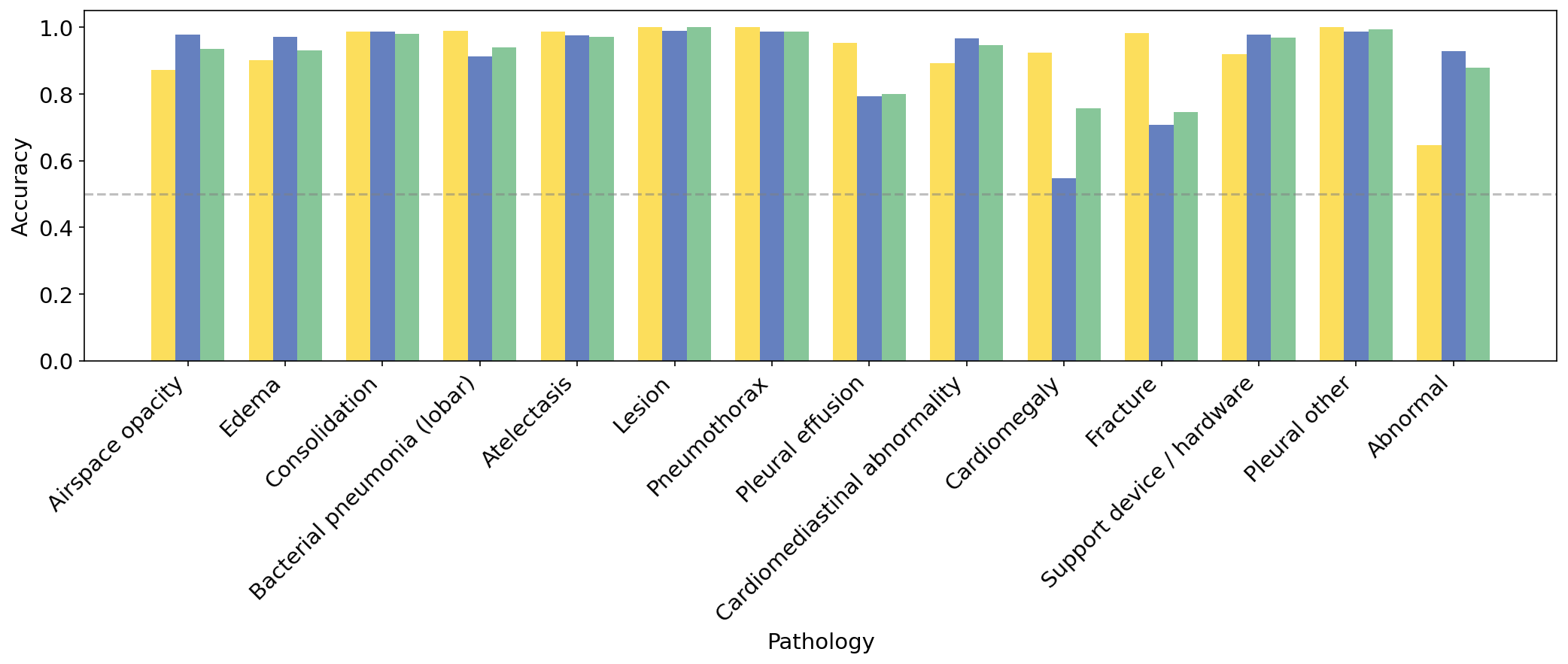}
    \caption{Accuracy of the AI predictor (yellow), non-guided human expert (blue) and AI-guided human expert (green) across pathologies. The AI predictor and AI-guided human outperform the non-guided human by the largest margin on pleural effusion, cardiomegaly and fracture.} 
    \label{fig:aivsplainvsguided}
\end{figure}

AI predictor interpretations were provided on a subset of the pathologies the radiologists annotated, therefore only these 14 pathologies were relevant to L2D and used during these experiments. Analysis of the dataset shows the AI predictor outperforms the human expert in some but not all cases, and AI guidance is helpful to the radiologists in some cases, but not in all (Fig. \ref{fig:aivsplainvsguided}).

The pooled radiologist annotations, CheXbert labeller output probabilities, and images were used for training, and images provided as model input for testing. 

\subsubsection{CheXpert and VinDr-CXR} To validate the findings on Collab-CXR, the experiments were repeated with two larger datasets, CheXpert \cite{Chexpert} and VinDr-CXR \cite{VinDr}. While these datasets have individual real human annotations, removing the need for pooling, they do not contain human annotations with AI assistance. Following the standard in the L2D literature, the guidance was generated through simulation. The performance of the AI predictor and the AI-guided human were simulated with a pathology-dependent method, with relative performance consistent with the real-world Collab-CXR dataset: flipping binary predictions from ground truth following \cite{l2dmozannar2020consistent}, with pathology-specific probability, to produce baseline accuracies within $\pm$ 0.9\% of each other, with the AI the best predictor on 46-64\% of pathologies, and AI-guided human performance in between AI and human on all but 7-14\% of cases.

\subsection{Models}
The ResNet model \cite{ResNet} for image encoding was pre-trained for 20 epochs on the National Institutes of Health Chest X-Ray Dataset (NIH-CXR) \cite{NIHCXR2017}.
The pathology-specific encoder was pre-trained on the NIH-CXR dataset with NIH pathology labels for 10 epochs, and was updated during rejector training.

A grid search was used in each experiment with Random Forest and XGBoost to tune the hyperparameters for the rejector model. For Random Forest: number of estimators (100-1000), maximum depth of estimator (0-30) and minimum number of samples required to split an internal node (2-10) were tuned. For XGBoost: number of estimators (200-1000), maximum depth of estimator (3-7), and learning rate (0.01-0.1) were tuned. The MLP rejector model consisted of three linear layers interspersed with non-linear activation layers. Training was for up to 100 epochs, with early stopping on validation loss after 10 epochs without improvement.

Experiments were performed for the same $n=20$ random seeds for each setup to ensure comparison across the same data splits. 

\section{Results}

\subsubsection{Baseline}
The baseline predictor and expert accuracies on Collab-CXR are AI-guided human 92.3\%,  AI alone 92.0\% and human alone 91.9\%.
\begin{table}[h]
\centering
\caption{Accuracy mean and standard deviation of different rejector models on Collab-CXR with different combinations of tabular and image features. Note that all rejector models outperform the baselines (0.923 AI-guided human, 0.920 AI alone, 0.919 human alone).}

\begin{tabular}{c|l|lll}
\hline
\multicolumn{1}{c|}{\multirow{2}{*}{Rej. Model}} & \multicolumn{1}{c|}{\multirow{2}{*}{Tabular input features}} & \multicolumn{3}{c}{Image features}                       \\
\multicolumn{1}{c|}{}  & \multicolumn{1}{c|}{}         & \multicolumn{1}{c}{Absent}      & \multicolumn{1}{c}{Present}     & \multicolumn{1}{c}{Label-specific}\\ \hline
\multirow{4}{*}{Rand. Forest}             &Exp. output, path., AI output  & 0.962±0.006 & 0.969±0.006 &   \multicolumn{1}{c}{\textemdash} \\ 
& Pathology, AI   output   & 0.962±0.007 & 0.968±0.007 &   \multicolumn{1}{c}{\textemdash}\\ 
& AI output & 0.955±0.007 & 0.964±0.005 &  \multicolumn{1}{c}{\textemdash} \\ 
& Pathology  & 0.974±0.005 & 0.961±0.007 &\multicolumn{1}{c}{\textemdash}\\
\hline
\multirow{4}{*}{XGBoost}                   &     Exp. output, path., AI output  & 0.959±0.010  & 0.961±0.007 &   \multicolumn{1}{c}{\textemdash}\\
& Pathology, AI output & 0.972±0.010  & 0.970±0.007  & \multicolumn{1}{c}{\textemdash} \\ 
& AI output & {0.964±0.006}  &0.961±0.009 &\multicolumn{1}{c}{\textemdash}\\
& Pathology   & 0.970±0.010   & 0.967±0.008 &\multicolumn{1}{c}{\textemdash} \\ \hline
\multirow{4}{*}{Simple MLP}                       
& Exp. output, path., AI output   & \textbf{0.975±0.005 }& 0.931±0.015 & \multicolumn{1}{c}{0.974±0.012}  \\
& Pathology, AI   output  & \textbf{0.975±0.005 }& 0.931±0.016 & \multicolumn{1}{c}{0.969±0.016} \\ 
 & AI output & 0.968±0.004 & 0.933±0.021 & \multicolumn{1}{c}{0.958±0.012}  \\
& Pathology   & \textbf{0.975±0.005} & 0.944±0.018 & \multicolumn{1}{c}{0.961±0.023}   \\
\hline
\multirow{4}{*}{Formal L2D}   
& Exp. output, path., AI output  & 0.944±0.023 & 0.921±0.010  & \multicolumn{1}{c}{0.920±0.007}  \\
& Pathology, AI   output   & 0.943±0.026 & 0.922±0.009 & \multicolumn{1}{c}{0.920±0.007}  \\  
& AI output  & 0.931±0.020  & 0.920±0.009  & \multicolumn{1}{c}{0.920±0.007}  \\
& Pathology   & 0.936±0.022 & 0.921±0.008 &\multicolumn{1}{c}{0.920±0.007}  \\
\hline                        
\end{tabular}
\label{tab:collabcxrresults}
\end{table}

\begin{table}[h]
\centering
\caption{Accuracy of baselines and best performing classic L2D and L2D-G models, across Collab-CXR, CheXpert and VinDr-CXR, with model architectures and input features listed (MLP = simple MLP, RF = Random Forest; P = pathology, AI = AI output, I = image features, LSI = label-specific image features).}
\begin{tabular}{l|ccc|ll}
\hline
\multicolumn{1}{c|}{} & \multicolumn{3}{c|}{Baseline} & \multicolumn{2}{c}{Best Model Performance} \\
\cline{2-6}
\multicolumn{1}{c|}{Dataset} & \multicolumn{1}{c}{AI} & \multicolumn{1}{c}{Human} & \multicolumn{1}{c|}{AI-guided Human} & \multicolumn{1}{c}{Classic L2D} & \multicolumn{1}{c}{L2D-G} \\
\hline
Collab-CXR & 0.920 & 0.919 & 0.923 & 0.920 (P, AI) & \textbf{0.975} (MLP: P $\pm$ AI) \\
CheXpert   & 0.910 & 0.902 & 0.901 & 0.950 (AI) & \textbf{0.988} (RF: AI, I) \\
VinDr-CXR      & 0.933 & 0.940 & 0.939 & 0.953 (P, AI, I) & \textbf{0.969} (MLP: P, AI, LSI) \\
\hline
\end{tabular}
\label{tab:headlineresults}
\end{table}

\subsubsection{L2D-G Performance}
L2D with Guidance has accuracy of up to 0.975$\pm$0.005 on Collab-CXR, while deferring 38.2$\pm$0.9\% of cases, compared to the best baseline accuracy of 0.923 (AI-guided human).
All rejector models for L2D with Guidance outperform all of these baselines with some input feature combinations (Table \ref{tab:collabcxrresults}). This suggests that L2D-G outperforms human-alone or AI-alone on this dataset, and that the rejector is able to capitalise on the different strengths and weaknesses of the predictor and the expert.

\subsubsection{Comparison Between Rejector Models}
\begin{figure}
    \centering
    \includegraphics[width=0.7\linewidth]{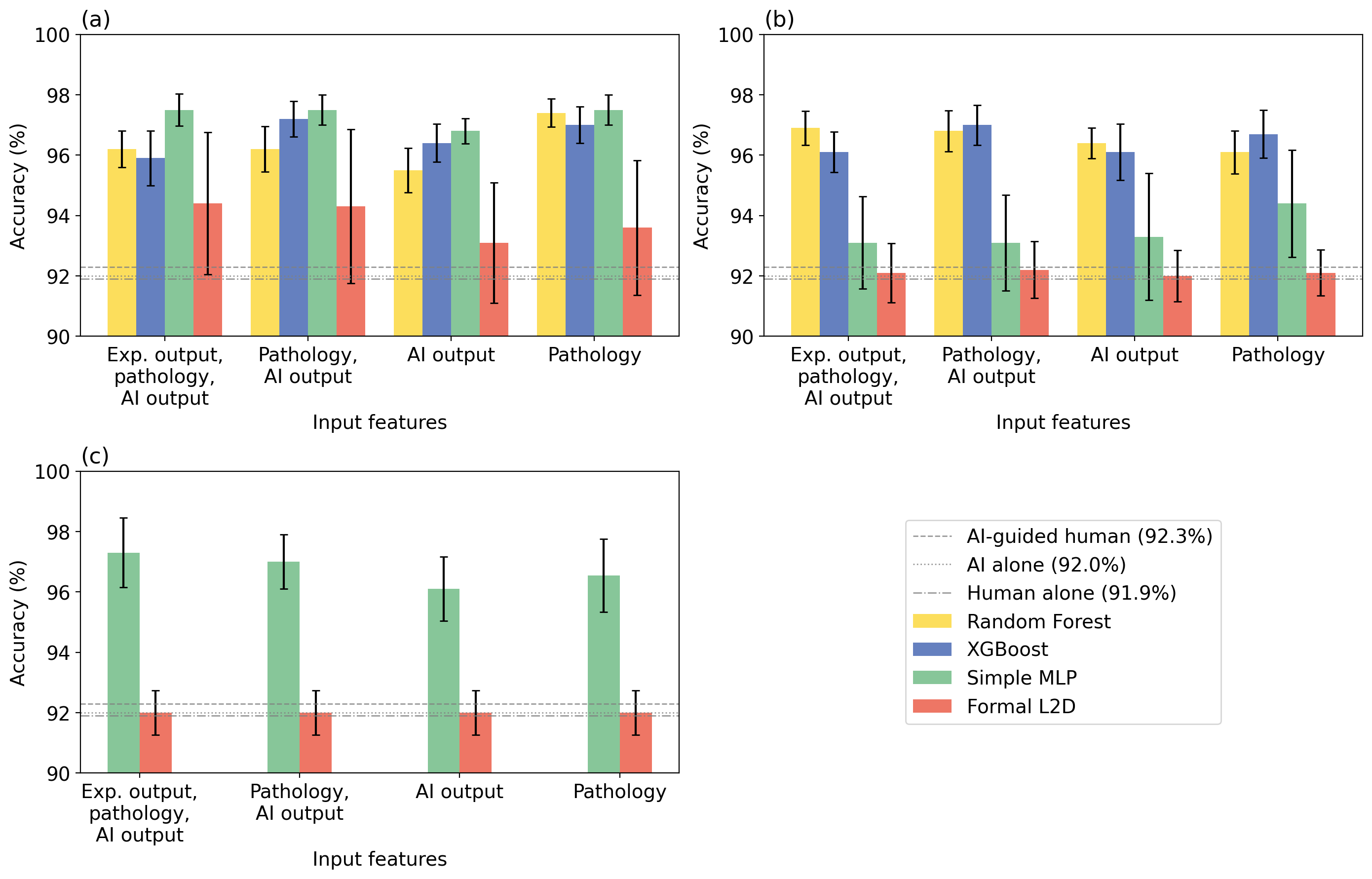}
    \caption{Performance of L2D with Guidance using (a) tabular features only, (b) image features, and (c) label-specific image embeddings, on Collab-CXR. Results are shown across different rejector models and input feature combinations, averaged over $n=20$ random seeds, and compared to baselines.
    Note that the y-axis is truncated for legibility.}
    \label{fig:multifigresults}
\end{figure}

Almost all models and input feature combinations outperform all baselines on Collab-CXR, the exceptions being the formal L2D rejector with image features or label-specific image embeddings included as input (Fig.\ref{fig:multifigresults}). The best performing model was the simple MLP on tabular data inputs.
These results show that with this dataset, the two-stage L2D-G model with the rejector using a formal surrogate loss function is outperformed by simpler rejector models such as an MLP with binary cross-entropy loss (paired $t$-test, $p=0.022$) (Table \ref{tab:collabcxrresults}). 
Additionally, for the best performing rejector model, the simple MLP, including more tabular data did not improve performance, including naive image features lowered performance, and including pathology-specific image embeddings did not improve performance above tabular data-alone. This is intuitively consistent, as the naive image embeddings do not reflect the relevance of different aspects of the image for diagnosing different pathologies - for instance cardiomegaly based on the size of the heart, compared to pleural effusion based on the opacity of the lungs. 

\begin{figure}
    \centering
    \includegraphics[width=0.75\linewidth]{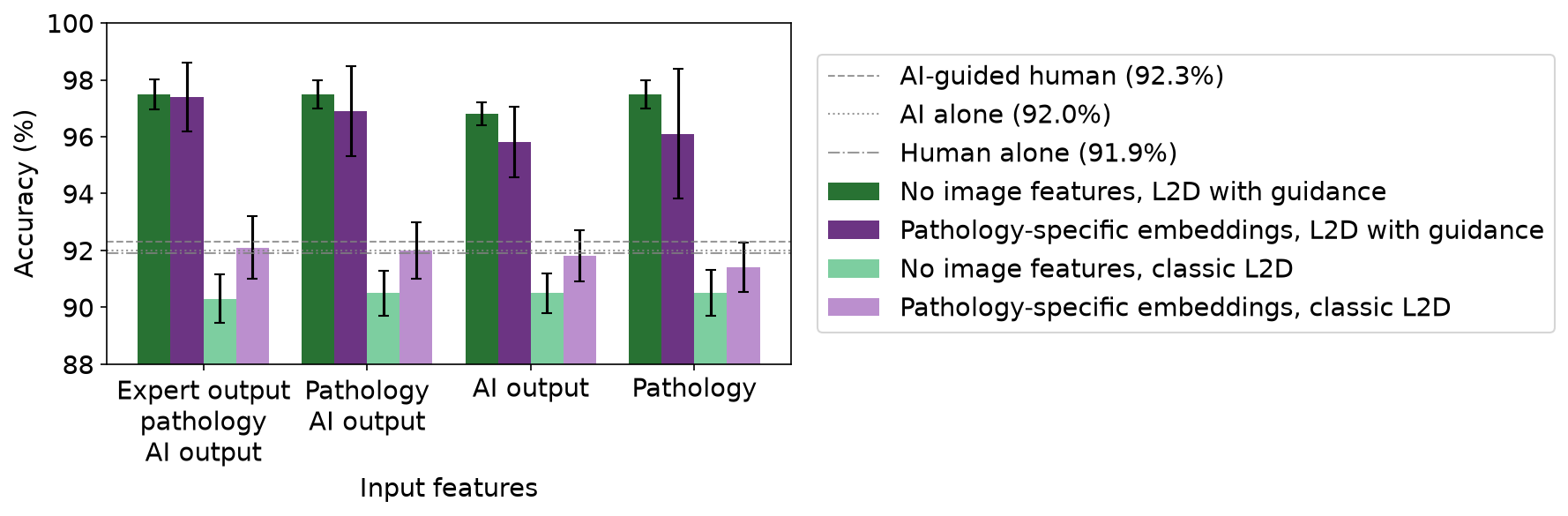}
    \caption{Accuracy of the L2D system with MLP rejector, comparing L2D-G to classic L2D, with and without pathology-specific image embeddings, on Collab-CXR. L2D-G outperforms baselines of AI-guided human, AI alone and human alone. Classic L2D performs worse than baselines on this data. 
    Note that the y-axis is truncated for legibility.}
    \label{fig:MLP_classic_vs_guided}
\end{figure}

\subsubsection{Classic L2D Performance}
This new formulation significantly outperforms classic L2D on Collab-CXR (paired $t$-test, $p<0.0001$) (Fig. \ref{fig:MLP_classic_vs_guided}). Classic L2D, with a single unguided expert, does not perform above the baselines. We hypothesise that with this moderately-sized dataset, without the option to defer to an AI-guided human, there is model overfitting and the system cannot learn useful features to defer between the AI and a single choice of expert. 

\subsubsection{Results on CheXpert and VinDr-CXR}
The same relative findings were demonstrated on the two augmented datasets. L2D-G had significantly higher accuracy compared to baseline, with the informal rejector models outperforming the formal L2D rejector. While classic L2D was able to outperform the baselines on these datasets, the best performance of L2D-G was considerably better than the best performance of classic L2D (Table \ref{tab:headlineresults}).

\section{Conclusion}

In this paper we proposed L2D with Guidance, a new formulation of Learning to Defer that provides optional AI guidance when deferring to a human expert. Validated on Collab-CXR, a publicly-available real-world medical imaging dataset, we achieved per-label accuracy above a baseline of 92.3\%. Notably, this performance is achieved with simpler loss functions compared to the formally defined L2D surrogate loss functions in existing literature. This finding is further reproduced on two further datasets, CheXpert and VinDr-CXR.

We also evaluated classic two-stage L2D for the first time on real-world human-annotated medical data and compare it to our new formulation. Our findings indicate weaker performance of two-stage L2D on these datasets, particularly on the real-world Collab-CXR dataset. 

L2D is a relatively new human-AI collaboration framework and while the theory under-pinning it has been evolving, to our knowledge this is the first paper evaluating it on a real-world medical imaging dataset with human expert annotations. This work is limited by the moderate size of the real-world dataset; as a next step, we are currently investigating the approach on private datasets across further imaging modalities. 
\begin{credits}
\subsubsection{Acknowledgments}
This work was supported by the Engineering and Physical Sciences Research Council. ES is funded by an EPSRC Doctoral Training Partnership [EP/W524311/1]. JS is funded by the EPSRC Center for Doctoral Training in Health Data Science [EP/S02428X/1]. AN acknowledges EPSRC Turing AI Fellowship: Ultra Sound Multi-Modal Video-based Human-Machine Collaboration [EP/X040186/1].

\subsubsection{Disclosure of Interests}
The authors have no competing interests to declare. \end{credits}
\newpage

\bibliographystyle{splncs04}
\bibliography{mybibliography}

@book{deployAIhealthcare,
author = {European Commission and Directorate-General for Health and Food Safety and EEIG and Open Evidence and PwC},
title = {Study on the deployment of {AI} in healthcare – Final report},
publisher = {Publications Office of the European Union},
year = {2025},
doi = {doi/10.2875/2169577}}

@inproceedings{l2dmozannar2020consistent,
  title={Consistent estimators for learning to defer to an expert},
  author={Mozannar, Hussein and Sontag, David},
  booktitle={ICML},
  pages={7076--7087},
  year={2020},
  organization={PMLR}
}

@article{cheXbert,
    author = {Akshay Smit and Saahil Jain and Pranav Rajpurkar and Anuj Pareek and Andrew Y. Ng and Matthew P. Lungren},
    title = {CheXbert: Combining Automatic Labelers and Expert Annotations for Accurate Radiology Report Labeling Using BERT},
    journal = {arXiv:2004.09167},
    year = {2020}
}

@article{CCXR,
  author  = {Moehring, A. and Kutwal, M. and Huang, R. and Banerjee, O. and Jacobi, A. and Eber, C. and Mendoza, D. and Chung, M. and Dayan, E. and Gupta, Y. and et al.},
  title   = {A dataset for understanding radiologist-artificial intelligence collaboration},
  journal = {Scientific Data},
  volume  = {12},
  number  = {1},
  year    = {2025},
  month   = {May},
  doi     = {10.1038/s41597-025-05054-0}
}

@article{Mao2023Defer,
  author  = {Mao, A. and Mohri, C. and Mohri, M. and Zhong, Y.},
  title   = {Two-stage learning to defer with multiple experts},
  journal = {Advances in NeurIPS},
  volume  = {36},
  pages   = {3578--3606},
  year    = {2023}
}

@inproceedings{NIHCXR2017,
  author    = {Wang, X. and Peng, Y. and Lu, L. and Lu, Z. and Bagheri, M. and Summers, R. M.},
  title     = {ChestX-ray8: Hospital-scale chest X-ray database and benchmarks on weakly-supervised classification and localization of common thorax diseases},
  booktitle = {Proceedings of the IEEE Conference on Computer Vision and Pattern Recognition},
  pages     = {2097--2106},
  year      = {2017}
}

@article{ResNet,
    author = {He, K. and Zhang, X. and   Ren, S. and Sun, J.},
    title = {Deep Residual Learning for Image Recognition},
    journal = {arXiv:1512.03385},
    year = {2015}
}

@article{XGBoost,
    author = {Tianqi Chen and Carlos Guestrin},
    title = {XGBoost: A Scalable Tree Boosting System},
    journal = {arXiv:1603.02754},
    year = {2016}
}

@article{RF, 
    title={Random forests}, 
    volume={45}, 
    DOI={10.1023/a:1010933404324}, 
    number={1}, 
    journal={Machine Learning}, 
    author={Breiman, Leo}, 
    year={2001}, 
    month={Oct}, 
    pages={5–32}}

@article{Rudolph_2024, title={Nonradiology health care professionals significantly benefit from {AI} assistance in emergency-related chest radiography interpretation}, volume={166}, DOI={10.1016/j.chest.2024.01.039}, number={1}, journal={CHEST}, author={Rudolph, Jan and Huemmer, Christian and Preuhs, Alexander and Buizza, Giulia and Hoppe, Boj F. and Dinkel, Julien and Koliogiannis, Vanessa and Fink, Nicola and Goller, Sophia S. and Schwarze, Vincent and et al.}, year={2024}, month={Jul}, pages={157–170}}

@article{Strong_2025, 
    author = {Strong, Josh and Sun, Emma and Rogers, Harry and Higham, Helen and Noble, Alison},
    title={Learning to Defer: A Survey},
    journal = {10.5281/zenodo.17843044},
    year = {2025}
}

@article{Chexpert, title={Chexpert: A large chest radiograph dataset with uncertainty labels and expert comparison}, volume={33}, DOI={10.1609/aaai.v33i01.3301590}, number={01}, journal={Proceedings of the AAAI Conference on Artificial Intelligence}, author={Irvin, Jeremy and Rajpurkar, Pranav and Ko, Michael and Yu, Yifan and Ciurea-Ilcus, Silviana and Chute, Chris and Marklund, Henrik and Haghgoo, Behzad and Ball, Robyn and Shpanskaya, Katie and et al.}, year={2019}, month={Jul}, pages={590–597}}

@article{VinDr,
  author = {Nguyen, Ha Quy and Pham, Hieu Huy and {Tuan Linh}, Le and Dao, Minh and Khanh, Lam},
  title = {{VinDr-CXR: An open dataset of chest X-rays with radiologist annotations}},
  journal = {{PhysioNet}},
  year = {2021},
  month = jun,
  note = {Version 1.0.0},
  doi = {10.13026/3akn-b287},
  url = {https://doi.org/10.13026/3akn-b287}
}

\end{document}